\documentclass[lettersize,journal]{IEEEtran}
\IEEEoverridecommandlockouts
\usepackage{cite}
\usepackage{amsmath,amssymb,amsfonts}
\newcommand\numberthis{\addtocounter{equation}{1}\tag{\theequation}}
\usepackage{algorithm}
\usepackage{algpseudocode}
\usepackage{hyperref}
\hypersetup{
    colorlinks=true,
    linkcolor=blue,
    filecolor=magenta,      
    urlcolor=blue,
    pdftitle={Overleaf Example},
    pdfpagemode=FullScreen,
    }
\usepackage{graphicx}
\usepackage{textcomp}
\usepackage{cite}
\usepackage{xcolor}
\usepackage{hyperref}
\usepackage[caption=false,font=normalsize,labelfont=sf,textfont=sf]{subfig}
\usepackage{mathtools}
\usepackage{bbm}
\usepackage{caption} 
\def\BibTeX{{\rm B\kern-.05em{\sc i\kern-.025em b}\kern-.08em
    T\kern-.1667em\lower.7ex\hbox{E}\kern-.125emX}}
\begin{document}
\newcommand{\floor}[1]{\lfloor #1 \rfloor}
\newcommand{\argmax}{\mathop{\mathrm{argmax}}\limits}

\title{Deep Unsupervised Learning Using Spike-Timing-Dependent Plasticity\\
}

\author{\IEEEauthorblockN{Sen Lu, Abhronil Sengupta} \\
\IEEEauthorblockA{\textit{School of Electrical Engineering and Computer Science} \\
\textit{The Pennsylvania State University}\\
University Park, PA 16802, USA \\
Email: \{senlu, sengupta\}@psu.edu}}


\maketitle

\begin{abstract}
Spike-Timing-Dependent Plasticity (STDP) is an unsupervised learning mechanism for Spiking Neural Networks (SNNs) that has received significant attention from the neuromorphic hardware community. However, scaling such local learning techniques to deeper networks and large-scale tasks has remained elusive. In this work, we investigate a Deep-STDP framework where a rate-based convolutional network, that can be deployed in a neuromorphic setting, is trained in tandem with pseudo-labels generated by the STDP clustering process on the network outputs. We achieve $24.56\%$ higher accuracy and $3.5\times$ faster convergence speed at iso-accuracy on a 10-class subset of the Tiny ImageNet dataset in contrast to a $k$-means clustering approach. \end{abstract}


\section{Introduction}
\IEEEPARstart{W}{ith} high-quality AI applications permeating our society and daily lives, unsupervised learning is gaining increased attention as the cost of procuring labeled data has been skyrocketing concurrently. The ever-more data-hungry machine learning models usually require a humongous amount of labeled data, sometimes requiring expert knowledge, to achieve state-of-the-art performance today. Since manual annotation requires a huge investment of resources, unsupervised learning is naturally emerging as the best alternative. 

One of the most prominent unsupervised learning methods is clustering. The main concept of clustering is to compress the input data (like images in the case of computer vision problems) into lower dimensions such that the low-dimensional features can be clustered into separable groups. The efficiency of the sample clustering process improves with better representations of the compressed features. Since the quality of features depends only on the dimension reduction algorithm, the design and choice of the clustering method are critical to the success of unsupervised learning. However, most real-world tasks are not easily represented as separable low-dimensional points. Earlier attempts include classical PCA reduction before clustering \cite{ding2004k}, while others attempt to augment more features with ``bags of features" \cite{csurka2004visual}; but mostly constrained to smaller tasks. Recent works like DeepCluster have explored scaling of unsupervised learning approaches by incorporating the $k$-means clustering algorithm with a standard Convolutional Neural Network (CNN)  architecture that can learn complex datasets such as ImageNet without any labels \cite{caron2018deep}. Some works have also proven that pre-training the network, even unsupervised, is beneficial to building the final model in terms of accuracy and convergence speed \cite{radford2015unsupervised, oord2018representation, radford2019language}.

The focus of this article, however, is on scaling unsupervised learning approaches in a relatively nascent, bio-plausible category of neural architectures - Spiking Neural Networks (SNNs). SNNs have been gaining momentum for empowering the next generation of edge intelligence platforms due to their significant power, energy, and latency advantages over conventional machine learning models \cite{sengupta2019going,davies2021advancing}. One of the traditional mechanisms of training SNNs is through Spike-Timing-Dependent Plasticity (STDP) where the model weights are updated locally based on firing patterns of connecting neurons inspired by biological measurements \cite{diehl2015unsupervised}. STDP based learning rules have been lucrative for the neuromorphic hardware community where various emerging nanoelectronic devices have been demonstrated to mimic STDP based learning rules through their intrinsic physics, thereby leading to compact and resource-efficient on-chip learning platforms \cite{saha2021intrinsic}. Recent works have also demonstrated that unsupervised STDP can serve as an energy-efficient hardware alternative to conventional clustering algorithms \cite{frady2020neuromorphic}. 

However, scaling STDP trained SNNs to deeper networks and complex tasks has remained a daunting task. Leveraging insights from hybrid approaches to unsupervised deep learning like DeepCluster \cite{caron2018deep}, we aim to address this missing gap to enable deep unsupervised learning for SNNs. Further, while techniques like DeepCluster have shown promise to enable unsupervised learning at scale, the impact of the choice of the clustering method on the learning capability and computational requirements remains unexplored.
The main contributions of the paper can therefore be summarized as follows:

\textbf{(i)} We propose a hybrid SNN-compatible unsupervised training approach for deep convolutional networks and demonstrate its performance on complex recognition tasks going beyond toy datasets like MNIST.

\textbf{(ii)} We demonstrate the efficacy of STDP enabled deep clustering of visual features over state-of-the-art $k$-means clustering approach and provide justification through empirical analysis by using statistical tools, namely Fisher Information Matrix Trace, to prove that STDP learns faster and more accurately.

\textbf{(iii)} We also provide preliminary computational cost estimate comparisons of the STDP enabled Deep Clustering framework against conventional clustering methods and demonstrate the potential of significant energy savings.

\section{Related Works}
\noindent \textbf{Deep Learning:} Unsupervised learning of deep neural networks is a widely studied area in the machine learning community \cite{bengio2012unsupervised, dike2018unsupervised}. It can be roughly categorized into two main methods, namely clustering and association. Among many clustering algorithms, $k$-means \cite{lloyd1982least}, or any variant of it \cite{krishna1999genetic, arthur2007k}, is the most well-known and widely used method that groups features according to its similarities. Its applications can be found in practice across different domains \cite{ng2006medical, kim2008recommender}. Other approaches focus on associations to learn data representations which are described by a set of parameters using architectures such as autoencoders \cite{rumelhart1986learning, hinton2006reducing} (where the data distribution is learnt by encoding features in latent space). 

In more recent works, such unsupervised learning methods have been applied to larger and more complex datasets \cite{caron2018deep}, making them applicable to more difficult problems. Further, recent advances in generative models have also provided opportunities at mapping unlabeled data to its underlying distribution, especially in the domain of image generation using Generative Adversarial Network (GAN) \cite{rombach2022high} with reconstruction loss directly \cite{bojanowski2017optimizing} or using the auto-encoded latent space \cite{kingma2013auto, bojanowski2017optimizing, masci2011stacked}. Dumoulin $et$ $al.$'s recent effort at combining GAN and auto-encoder has demonstrated even better performance \cite{bojanowski2017optimizing}.

\noindent \textbf{Bio-Plausible Learning}: Visual pattern recognition is also of great interest in the neuromorphic community \cite{diehl2015fast, neftci2014event}. In addition to standard supervised vision tasks, SNNs offer a unique solution to unsupervised learning - the STDP learning method \cite{diehl2015unsupervised}. In this scheme, the neural weight updates depend only on the temporal correlation between spikes without any guiding signals, which makes it essentially unsupervised. While it offers a bio-plausible solution, it is rarely used beyond MNIST-level tasks\cite{diehl2015unsupervised, lee2018pretrain, liu2019stdpLearning} and primarily used for single-layered networks. Going beyond conventional STDP based learning, Lee \textit{et al.} \cite{lee2018pretrain} proposed an STDP-based pre-training scheme for deep networks that greedily trained the convolutional layers' weights, locally using STDP, one layer at a time but limited only to MNIST. Similarly, in Ferre \textit{et al.}'s work \cite{ferre2018unsupervised}, the convolutional layers were trained on CIFAR10 and STL-10 with simplified STDP, but the layers were also trained individually with complex mechanisms. 
Beyond the STDP framework, some studies draw inspiration from alternative biological mechanisms such as local learning (DECOLLE) \cite{decolle_kaiser2020synaptic}, equilibrium-state-based learning (Equilibrium Propagation) \cite{scellier2017equilibrium, martin2021eqspike, bal2022sequence}, Implicit Differentiation \cite{bai2019deep, xiao2021training, bal2023spikingbert}, among others to achieve bio-plausible learning with much less dependence on the gradient. However, most works involve significantly more complex hardware implementation than STDP based learning approaches. For instance, DECOLLE requires the local computation of loss and backpropagation of errors at each layer, thereby introducing additional overhead. Similarly, Equilibrium Propagation requires the determination of rate of change of the spiking rate of the neurons to perform local weight updates \cite{martin2021eqspike}.

Our work explores a hybrid algorithm design based on a merger of the above two approaches. Our proposed framework provides a global training signal for the CNN using a straightforward and end-to-end STDP-based SNN implementation. We demonstrate significant accuracy improvement and computation savings for VGG-15 architecture on the Tiny ImageNet dataset in contrast to state-of-the-art deep clustering approaches.

\section{Preliminaries}

\subsection{Deep Clustering with k-means Algorithm}
Deep Clustering \cite{caron2018deep} enabled unsupervised training of visual features primarily relies on the ability of clustering algorithms like the $k$-means to group together similar data points. $k$-means is a popular unsupervised algorithm for separating data points into distinct clusters. Given a user-specified value of $k$, the algorithm will find $k$ clusters such that each data point is assigned to its nearest cluster. The vanilla implementation of the $k$-means algorithm iteratively calculates the Euclidean distance between points for comparison and updates the cluster centroids to fit the given distribution. 

Deep Clustering utilizes the traditional CNN architecture to obtain the features to be used for clustering. The reason behind this feature reduction choice hinges upon the fact that a randomly initialized and untrained CNN outperforms a simple multilayer perceptron network by a considerable margin \cite{noroozi2016unsupervised}. Driven by this observation, the main idea behind this framework is to bootstrap the better-than-chance signal to teach the network and learn the features. This teaching signal is transformed into a `pseudo-label' so that the network can learn from it. The `pseudo-labels' which may or may not be the same as the ground truth labels reflect the direction that the network weights should be updated. By doing so, the feature extraction layers may become slightly better at recognizing certain features and thereby producing more representative features. The improved features can ideally be more separable, thereby generating higher quality `pseudo-labels'. By repeating this process iteratively, the CNN should ideally converge by learning the `pseudo-labels' \cite{caron2018deep}. 

Note that the CNN layers used for feature-reduction purposes can be converted into SNN layers with various methods as shown in many recent studies \cite{midya2019artificial, sengupta2019going, lu2020exploring, lu2022neuroevolution, gao2023high}, or trained from scratch using backpropagation through time (BPTT) \cite{bellec2018long,Rathi2020DIETSNNDI} which opens up the potential for adopting the entire feature-reduction in a low-power neuromorphic setting. In this work, we therefore do not focus on the CNN-SNN conversion and train it by backpropagation without unrolling through time.




\subsection{STDP Enabled Neuromorphic Clustering}
STDP is an unsupervised learning mechanism that learns or unlearns neurons' synaptic connections based on spike timings \cite{caporale2008spike}. In particular, the synaptic connection is strengthened when the post-synaptic neuron fires after the pre-synaptic neuron, and the connection is weakened if the post-synaptic neuron fires before the pre-synaptic neuron. The intuition behind STDP follows Hebbian learning philosophy where neurons that are activated together and sequentially are more spatio-temporally correlated and thus form a pattern, and vice versa. This learning rule enables the encoding of complex input distributions temporally without the need for guiding signals such as the label. The weights of the neuronal synapses are updated based on spike timings \cite{diehl2015unsupervised} as follows:
\begin{equation}
\Delta w = \begin{cases}
    A_+e^{-\frac{\Delta t}{\beta_+}}, &\text{if}\ \Delta t > 0\\
    -A_-e^{\frac{\Delta t}{\beta_-}}, &\text{if}\ \Delta t < 0
\end{cases}
\end{equation}
where, $w$ is the weight, $A_{+/-}$ are the learning rates, $\Delta t$ is the exact time difference between post-neuron and pre-neuron firing and $\beta_{+/-}$ are the time-constants for the learning windows. In practical implementations, the exact spike timing is usually replaced with a spike trace (see Section IV-B) that decays over time to reduce memory storage for STDP implementation \cite{Hazan_2018}.

\begin{figure}[t]
  \centering
  \includegraphics[width=0.2 \textwidth]{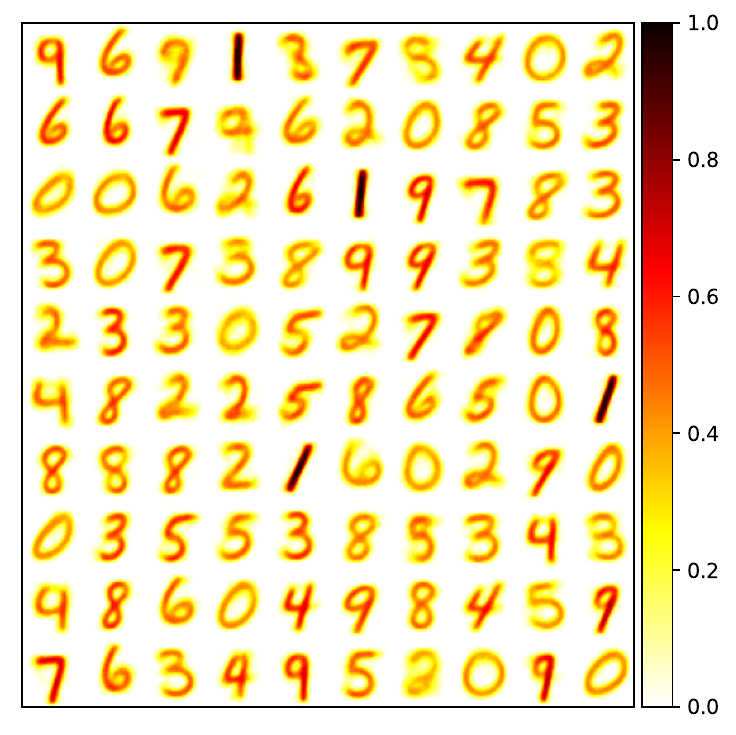}
  \caption{STDP learns generic features of input patterns (MNIST dataset) in the excitatory synapses of the Winner-Take-All network. Each neuron represents a cluster and its learnt weights represent the corresponding cluster centroid.}
  \label{fig:mnist}
    \vspace{-1em}
\end{figure}

STDP training is predominantly explored in Winner-Take-All networks in literature which consists of an excitatory layer of neurons with recurrent inhibitory connections \cite{diehl2015unsupervised} (see ``STDP Enabled SNN for Clustering" sub-panel in Fig. \ref{fig:struct}). Such connections create a mechanism called `lateral inhibition' where activated neurons inhibit other neurons' activities and therefore assist the activated neurons to accentuate the learning process of its weights. To prevent any neuron from dominating the firing pattern, the second key mechanism is `homeostasis' which balances the overall activities of the neurons. Homeostasis prevents neurons from runaway excitation or total quiescence. One popular way to achieve this is through adaptive and decaying thresholding in which after every firing event, the firing threshold increases such that the firing neuron requires higher membrane potential to fire again in the future. Consequently, this will provide opportunities for other neurons in the network to fire and learn the synaptic weights. The critical balance of these two mechanisms ensures stable learning of the SNN. Fig. \ref{fig:mnist} shows an example of STDP-trained weights of the excitatory neuron layer of an SNN where representative digit shapes are learnt without any label information for the MNIST dataset \cite{deng2012mnist}. Each neuron in the network represents a cluster. By running inferences on the STDP network, we can cluster the inputs according to their corresponding most activated neuron. The learnt weights of each neuron is equivalent to the centroid of the cluster represented by that neuron.   

\section{Methods}

\subsection{Proposed Deep-STDP Framework}
\begin{figure}[h]
\vspace{-1em}
  \centering
  \includegraphics[trim=150pt 0pt 100pt 0pt,clip, scale=0.37]{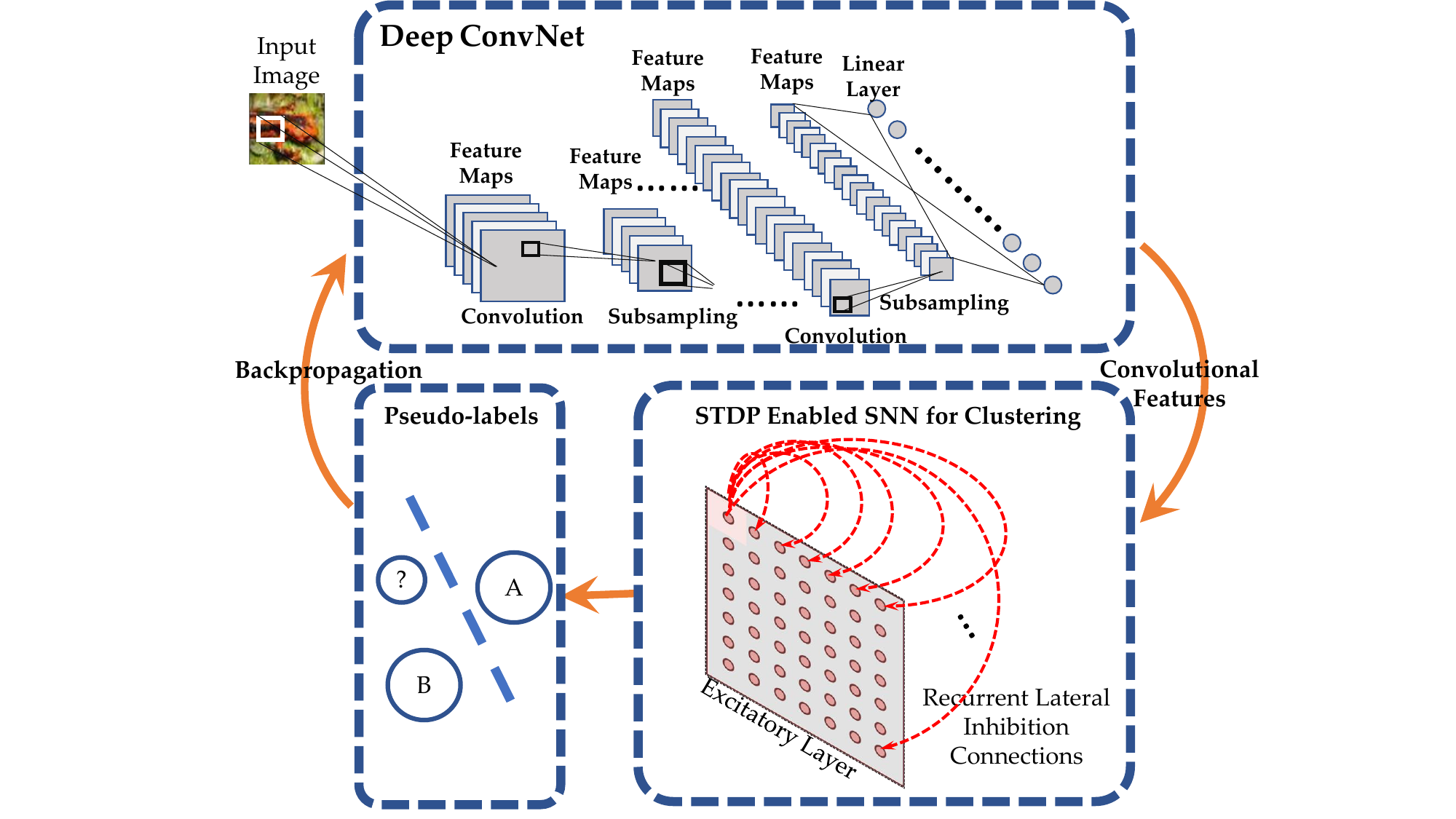}
  \caption{Overall structure of Deep-STDP: The ConvNet compresses the input images to a lower-dimensional feature vector which is mapped by STDP clustering to a pseudo-label. The ConvNet is subsequently trained through backpropagation using the pseudo-labels.}
  \label{fig:struct}
  \vspace{-1em}
\end{figure}

As mentioned previously, the convolutional layers of the network compress the input images to a lower dimensional feature space as a one-dimensional vector. In abstract terms, the framework solves the following optimization problem \cite{caron2018deep}:
\vspace{-1em}
\begin{align*}
    \min _{w \in \mathbb{R}^{d \times k}} \frac{1}{N} \sum_{n=1}^{N} \min _{y_n \in {\{0,1\}}^k}||f_\theta (img_n) - w_{y_n} ||_1 \numberthis\\
    \text{ such that }y_n^\intercal 1_k = 1 
\end{align*}
where, $N$ is the total number of training samples, $y_n$ is the $n$-th optimal neuron assignment encoded as a one-hot vector, $f_\theta$ is the ConvNet forward pass output parameterized by its weights $\theta$, $img_n$ is the $n$-th input sample, $w_{y_n}$ is the STDP-learnt synaptic weight map of the most activated neuron, $d$ is the feature dimension of the ConvNet output and $k$ is the number of neurons/clusters in the network. By minimizing the difference between the weights of the neurons and the patterns of the features, we can obtain an SNN that generates optimal assignments of $y_{n}$ parameterized by weights $w$, which act as the pseudo-labels for our algorithm.

With the pseudo-labels, the network training can be accomplished through the standard minimization problem of network loss which can be described by:
\begin{equation}
    \min _{\rho, \theta} \frac{1}{N} \sum_{n=1}^{N} \mathcal{L} (g_\rho (f_\theta (img_n)), y^*_n )
\end{equation}
where, $\theta, \rho$ are parameters of the ConvNet $f_\theta (\cdot)$ and classifier $g_\rho (\cdot)$ respectively, $\mathcal{L(\cdot)}$ is the loss function,  $img_n$ again is the $n$-th image input, $y^*_n$ is the $n$-th optimal pseudo-label for this iteration. 

However, SNNs only accept discrete spikes as input and therefore the ConvNet feature outputs in floating-point representation (after appropriate pre-processing like PCA reduction and $l_2$-normalization \cite{caron2018deep}) are subsequently rate encoded by a Poisson spike train generator, where the feature values are used as the Poisson distribution rate and sampled from the respective distribution. 
At the end of the pseudo-label assignment, the STDP enabled SNN resets for the next iteration. This is intuitive since after the ConvNet weight update process, the feature distribution gets shifted and hence a new set of neuron/cluster weights should be learnt by the STDP framework. Algorithms \ref{alg:deepSTDP}-\ref{alg:pseudo_label} describe the overall structure of the proposed Deep-STDP framework shown in Fig. \ref{fig:struct}.

\algrenewcommand\algorithmiccomment[1]{\hfill\textcolor{cyan}{// #1}}
\begin{algorithm}
\caption{DeepSTDP}\label{alg:deepSTDP}
\begin{algorithmic}[1]
    \Require $k = 100$ \Comment{\# of expected clusters$\times$10}
    \Require $epochs \geq 200$ \Comment{training epochs limit}
    \State \text{Initialize $ConvNet$ randomly}
    \For{$1$ to $epochs$}  
        \State $ConvNet  \Rightarrow $ \text{Remove last layer}
        \State \text{Initialize $DatasetX$}  \Comment{Training images} 
        \For{$img \in DatasetX$}                    
            \State Run $ConvNet$ on $img$ $\Rightarrow features $ \Comment{Inference}
        \EndFor
        \State  \text{Rate encodes $features$}$ \Rightarrow featureSet$\
        \State \Call{STDP}{$featureSet$} $ \Rightarrow labels_{pseudo} $ \Comment{Train the SNN and generate pseudo-labels using Algo. \ref{alg:pseudo_label}}
        
        \State $ConvNet  \Rightarrow $ \text{Adds back last layer}
        \State Train $ConvNet$ on $DatasetX$ and $labels_{pseudo}$ \Comment{Standard learning through backpropagation}
    \EndFor

\end{algorithmic}
\end{algorithm}
\algrenewcommand\algorithmiccomment[1]{\hfill\textcolor{cyan}{// #1}}
\begin{algorithm}
\caption{STDP}\label{alg:pseudo_label}
\begin{algorithmic}[1]
    \Require $k = 100$ \Comment{\# of neurons}
    \Require Table \ref{table:param} parameters \Comment{See Table I for details}
    \Require $timesteps = 100$ \Comment{Simulation steps}
    \Require $featureSet$ \Comment{Features to be trained on}
    \State Initialize $w_+, w_-$ \text{randomly}
    \State Initialize $k$ neurons with resting potential $V_{rest}$
    \State Initialize $w_{inh}$ \text{to -1}  \Comment{Fixed inhibitory weights}
    \State $l = 0$ \Comment{Refractory counter initialized for all neurons}
    \State $\epsilon = 0$ \Comment{No extra threshold initially}
    \For{$(feature \in featureSet)$, $(t \Rightarrow 1$ to $timesteps)$} 
        \State \text{Encode }$ feature \Rightarrow s^{pre}_{+}, s^{pre}_{-} $
        \State Resets $\tau^{pre}_{+} , \tau^{pre}_{-} $ to $\tau_o$ based on spikes 
        \State $\tau^{pre}_{+} = $  $exp(\frac{1}{\tau_{decay}}) \tau^{pre}_+ $ 
        \State $ \tau^{pre}_{-} = $  $exp(\frac{1}{\tau_{decay}}) \tau^{pre}_- $ 
        \State $\tau^{post} =$ $exp(\frac{1}{\tau_{decay}}) \tau^{post} $  \Comment{Decay spike traces}
        \State $\epsilon = exp(\frac{1}{\epsilon_{decay}}) \epsilon$ \Comment{Decay adaptive threshold}
        \State Decay potential using Eq. \ref{eq:v_decay} $\Rightarrow V^{t}_{exc}$
        \State $l = max(l - 1, 0)$ \Comment{Count down refractory period}
        
        \For{Neurons with $(l==0)$} \Comment{Non-refractory}
            \State $V^{t}_{exc} \gets $\text{Update using Eq. \ref{eq:membrane}}
            \State Fire spikes when $(V^t_{exc} > V_{thr} + \epsilon) \Rightarrow s^{post}$ 
            \For{Neurons with ($V^t_{exc} > V_{thr} + \epsilon$)}
                \State $l = L$ \Comment{Reset refractory period}
                \State $\epsilon = \epsilon + \alpha$ \Comment{Add adaptive threshold $\alpha$}
                \State Resets $\tau^{post}$ to $\tau_o$ based on spikes 
        
                
                
            \EndFor
            
        \EndFor
        \State Update using Eq. \ref{eq:weights_pre} and \ref{eq:weights_post}  $\Rightarrow \Delta w_{+}, \Delta w_{-}$ 
        \State $V^{t}_{exc} $ excluding itself $\gets \text{$s^{post}$} \cdot \text{$w_{inh}$} $ \Comment{Lateral inhibition} 
        
 
\EndFor
    \State \Return $labels_{pseudo} \Rightarrow$ most activated excitatory neuron IDs
\end{algorithmic}

\end{algorithm}

\subsection{STDP Enabled SNN for Clustering}
Clustering in the SNN is mediated through the temporal dynamics of Leaky-Integrate-Fire neurons in the excitatory layer. In the absence of any spiking inputs, the membrane potential of neurons in the excitatory layer is represented by $V_{exc}$ at timestep $t$, or simply $V_{exc}^t$. It initializes with $V_{exc}^{t=0} = V_{rest}$ and decays as,
\begin{equation}
V_{exc}^t = V_{rest} +  exp(\frac{1}{V_{decay}}) (V_{exc}^{t-1}-V_{rest}) 
\label{eq:v_decay}
\end{equation}
where, $V_{rest}$ is the resting potential and $V_{decay}$ is the potential decay constant.

Prior works \cite{diehl2015unsupervised} on using SNNs for clustering have mainly dealt with simple datasets without negative-valued features. This is in compliance with the nature of STDP learning for positive valued spikes. However, in our scenario, we consider negative valued spiking inputs as well in order to rate encode the negative features provided as output of the ConvNet. In order to enable STDP learning for negative inputs, we decompose the weight map into positive and negative components to learn positive and negative spike patterns respectively. Therefore, in presence of spikes, the excitatory layer's neuron membrane potential dynamics is updated as,
\begin{equation}
V_{exc}^t \gets \text{$s^{pre}_{+}$} \cdot \text{$w_{+}$} + \text{$s^{pre}_{-}$} \cdot \text{$w_{-}$}
\label{eq:membrane}
\end{equation}
where, the membrane potential is denoted by $V^t_{exc}$ at timestep $t$, and the input spikes and pre-synaptic weights are represented by $s^{pre}$ and $w$ respectively (with their positive and negative counterparts).  It is worth mentioning here that pre-neurons refer to the input neurons and post-neurons refer to the excitatory layer neurons since the synapses joining them are learnt by STDP. 

Further, there is a refractory period $L$ parameter for every neuron which will only allow execution of Eq. \ref{eq:v_decay} and \ref{eq:membrane} if the refractory counter, $l$, equals `0'. A spike will be generated when the membrane potential at the current timestep is greater than the membrane threshold:
\begin{equation}
    s= \begin{cases} 
        1 & \text{if } (V^t_{exc} > V_{thr} + \epsilon)\text{ and }(l=0)\\
        0 & \text{otherwise}
        \end{cases}
        \label{eq:spiking}
\end{equation}
where, $V_{thr}$ is the membrane threshold to fire a spike, $\epsilon$ is the adaptive threshold parameter, $l$ is the refractory period counter which is reset to $L$ upon a firing event and decays by 1 otherwise (thereby preventing neurons from firing for $L$ timesteps after a spike). $V^t_{exc}$ resets to $V_{reset}$ after firing a spike. The adaptive threshold parameter acts as a balancer to prevent any neuron from being over-active (homeostasis) and is incremented by parameter $\alpha$ upon a firing event and otherwise decays exponentially at every timestep similar to Eq. \ref{eq:v_decay}: $exp(\frac{1}{\epsilon_{decay}}) \epsilon$. Every spike generated by a post-neuron triggers a membrane potential decrement by an amount $w_{inh}$ for all the other neurons except itself.

In the context of our implementation, we used the spike trace $\tau$ to represent the temporal distance between two spikes. The spike trace value peaks at its firing to $\tau_{o}$ and exponentially decay as time lapses: $exp(\frac{1}{\tau_{decay}}) \tau$. The weight updates are similarly separated into positive and negative parts.\\
\textbf{Pre-synaptic update:}
\begin{equation}
\begin{array}{l@{}l}
    \Delta w_{+} &= -\eta^{pre} (s^{pre}_{+} * \tau^{post})\\
    \Delta w_{-} &= \eta^{pre} (s^{pre}_{-} * \tau^{post})
\end{array}
\label{eq:weights_pre}
\end{equation}
\textbf{Post-synaptic update:}
\begin{equation}
\begin{array}{l@{}l}
    \Delta w_{+} &= \eta^{post} (\tau^{pre}_{+} * s^{post})\\
    \Delta w_{-} &= \eta^{post} (\tau^{pre}_{-} * s^{post})
\end{array}
\label{eq:weights_post}
\end{equation}
where, $\Delta w$ are the weight updates, $\eta^{pre}, \eta^{post}$ are the learning rates for pre- and post-synaptic updates respectively, $\tau$ is the spike trace, and $s$ is the spiking pattern. Superscript $(^{pre})$, $(^{post})$ indicates whether the trace or spike is from pre- or post-synaptic neuron respectively, and the subscript $(_{+})$, $(_{-})$ indicates whether the operation is for positive or negative input spikes. Note that the negative $s^{pre}_{-}$ can be flipped easily by the distributive property of matrix multiplication. 

\section{Experiments and Results}

\subsection{Datasets and Implementation}
The proposed method was evaluated on the Tiny ImageNet dataset, which is a center-cropped subset of the large-scale ImageNet dataset \cite{deng2009imagenet}. Unlike the ImageNet 2012 dataset, which contains 1000 object categories, the Tiny ImageNet dataset comprises of only 200 categories. Due to computation constraints, we selected the first 10 classes from the Tiny ImageNet dataset by the naming order and considered both the training and testing sets for those corresponding classes in this work. All images were normalized to zero mean and unit variance and shuffled to avoid any bias. We chose VGG15 as the baseline network architecture with randomly initialized weights. Simulations were conducted using the PyTorch machine learning library and a modified version of the BindsNet toolbox \cite{Hazan_2018} as the base platform for the experiments.  The results reported for the DeepCluster framework \cite{caron2018deep} were obtained without any modification to the open-source codebase associated with the work, and its hyperparameters were unchanged unless mentioned in this work. The ConvNet learning rate was set to $1e-2$ and the number of clusters was set to 10 times the number of classes (recommended as optimal in Ref. \cite{caron2018deep} and also found optimal in the Deep-STDP framework). The training was performed for 200 epochs. All results obtained were run on 2 GTX 2080Ti GPUs and the associated hyper-parameters used for the Deep-STDP framework can be found in Table \ref{table:param}.

\begin{table}[h!]
\centering
\begin{tabular}{|c|c|c|}
\hline
\textbf{Parameter} & \textbf{Value} & \textbf{Description}\\
\hline
$k$         & 100 & Number of neurons in the network\\
$V_{rest}$  & -65 & Resting membrane potential\\
$V_{reset}$ & -60 & Resetting membrane potential\\
$V_{decay}$ & 20 & Membrane potential decay constant \\
$V_{thr}$       & -52 & Threshold potential to fire a spike\\
$L$         & 5  & Refractory period\\
$\tau_o$    & 1.0  & Trace value when a spike fires\\
$\tau_{decay}$& 100 & Trace decay constant\\
$\alpha$    & 0.45 & Adaptive threshold increment\\
$\epsilon_{decay}$& $1e7$ & Adaptive threshold decay constant\\
$\eta _{pre}$& $1e-3$ & Pre-synaptic learning rate\\
$\eta _{post}$& $1e-6$ & Post-synaptic learning rate\\
$w_{inh}$&  $-1$ & Fixed inhibitory recurrent connection weights\\
$T$         & 400  & SNN simulation duration\\
\hline
\end{tabular}
\caption{Hyper-parameters for STDP training}
\label{table:param}
\vspace{-2em}
\end{table}

Numerous cluster re-assignment frequencies were explored and `1' (`2') was found to be the optimal for Deep-STDP (DeepCluster), i.e. the pseudo-labels were generated by passing the entire dataset once (twice) every epoch. Note that this frequency represents the number of dataset iterations per epoch. Following the evaluation method proposed by Zhang \textit{et. al} \cite{zhang2016colorful}, we froze all network parameters and trained a linear layer at the output to evaluate the efficiency of the model to capture the distribution of images in the training set as well as its usage as a pre-trained model for general use cases. We fixed the random seeds in each experiment such that the clustering process is deterministic for a particular run. To prevent loss in generality, all accuracy results reported here represent the average value over 5 independent runs with different sets of random seeds.

\subsection{Evaluation Metrics}
\subsubsection{Fisher Information}
The Fisher information (FI) quantitatively measures the amount of information retained in a statistical model after being trained on a given data distribution \cite{amari2000methods}. Many prior works have used this metric to measure different aspects of deep learning models including SNN models \cite{karakida2019universal, kim2022exploring}. Unlike prior works, we use pseudo-labels to generate FI instead of ground-truth labels. FI reflects the impact of weight changes on the ConvNet output. If the FI of model parameters is small, we can conclude that the model's learning efficiency is poor since the weights can be pruned without affecting the output, and vice versa. Therefore, this metric implicitly measures the quality of the pseudo-labels.

Let us consider that the network tries to learn $y$ from a distribution $p$ parametrized by a set of weights $\theta$. Given samples $x$, the posterior distribution is $p_{\theta}(y|x)$. 
The Fisher information matrix (FIM) is defined as:
\begin{equation}
    F=\mathbb{E}_{x\sim X} \mathbb{E}_{y\sim p_\theta(y|x)} [\nabla_\theta \log p_\theta(y|x) \nabla_\theta \log p_\theta(y|x)^T]
\end{equation}
where, $X$ is the empirical distribution of the actual dataset. However, the exact FIM is usually too large to be computed directly and therefore the value is usually approximated by its trace, which is given by:

\begin{equation}
    \mathrm{Tr}(F)=\mathbb{E}_{x\sim X} \mathbb{E}_{y\sim p_\theta(y|x)} [||\nabla_\theta \log p_\theta(y|x)||^2]
\end{equation}
in which the expectations can be replaced by the averaged observation from the dataset of $N$ samples:
\begin{equation}
\mathrm{Tr}(F) = \frac{1}{N} \sum_{k=1}^N ||\nabla_\theta \log p_\theta(y|x)||^2_2
\end{equation}
where, $\mathrm{Tr}(F)$ is the trace of FIM, $\nabla$ is the partial derivative operator. We follow the same implementation as the algorithm specified in Ref. \cite{kim2022exploring}.

\subsubsection{Normalized Mutual Information }
Further, following the Deep Clustering work \cite{caron2018deep}, we also measured the Normalized Mutual Information (NMI) metrics to evaluate mutual information between two consecutive assignments of the STDP-enabled SNN, given by Eq. \ref{eq:nmi}.
\begin{equation}
\text{NMI}(y^p,y^{p-1}) = \frac{I(y^p;y^{p-1})}{\sqrt{[H(y^p)H(y^{p-1})]}}
\label{eq:nmi}
\end{equation}
where, $y^p,y^{p-1}$ are label assignments for epoch $p-1$ and $p$ respectively, $I(\cdot)$ is the mutual information function, and $H(\cdot)$ is the entropy function. 
Since the assignments $y^p, y^{p-1}$ are consecutive and are generated from the same inputs, a high NMI value indicates a high correlation between the two sets of assignments as well as stable assignments of the pseudo-labels. 

\subsection{Performance Evaluation}
\vspace{-1em}
\begin{figure}[htp]
    \centering
    \includegraphics[scale=0.4]{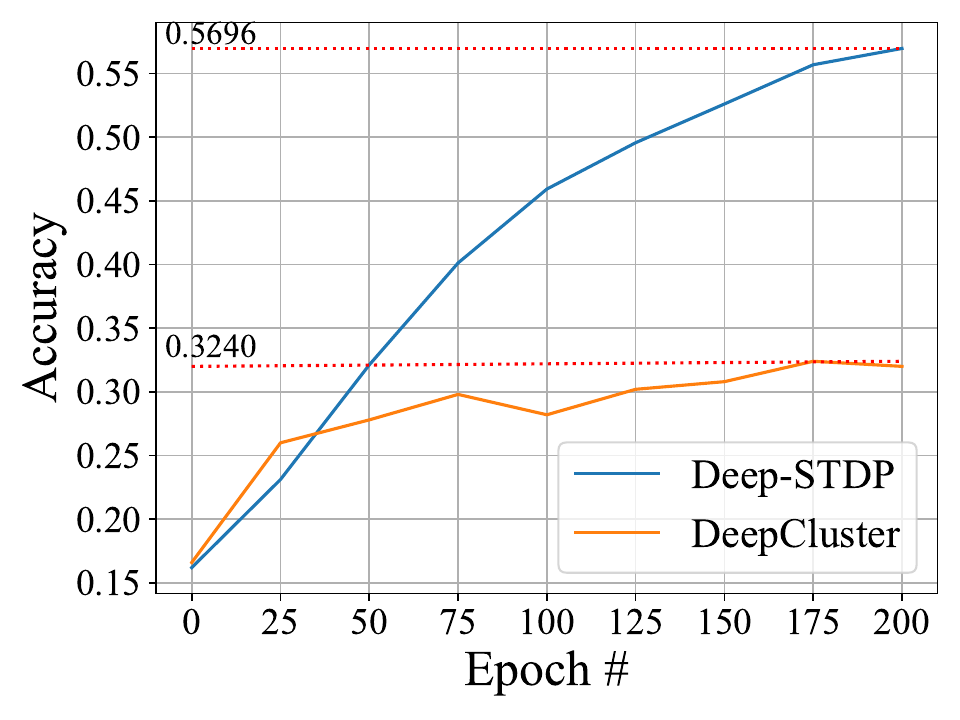}
    
\caption{Average accuracy over 5 independent runs of Deep-STDP and DeepClustering frameworks for the 10-class subset of Tiny ImageNet dataset.\label{fig:acc}}
\end{figure}
\vspace{-1em}
\begin{figure}[htp]
    \centering
    \includegraphics[scale=0.4]{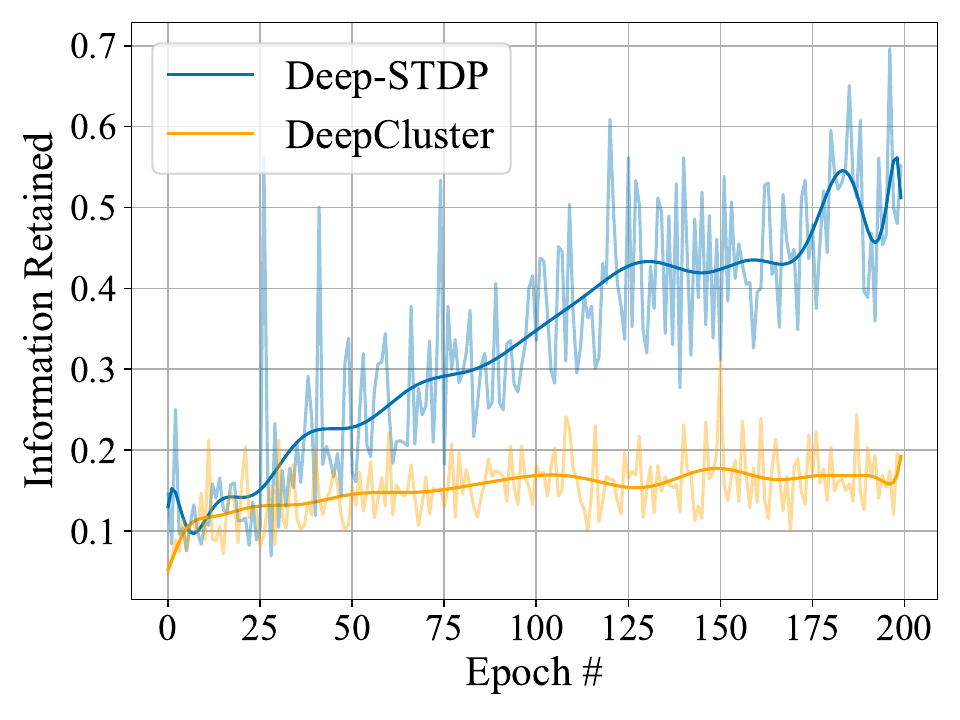}
\caption{Fisher information metric comparison: Deep-STDP retains more information over epochs\label{fig:fisher}. Pale curves are the actual data while darker curves are the smoothed plots.}
\end{figure}
Fig. \ref{fig:acc} demonstrates that Deep-STDP based unsupervised feature learning significantly outperforms DeepCluster approach based on $k$-means clustering. The superior quality of pseudo-labels generated by Deep-STDP is also explained empirically by the FIM trace variation over the learning epochs (see Fig. \ref{fig:fisher}). While both algorithms perform similarly during the initial stages, the accuracy and FIM trace start improving significantly for the Deep-STDP approach over subsequent epochs. Performance evaluation metrics (NMI, FIM and Accuracy) for the two approaches at the end of the training process are tabulated in Table \ref{table: results}.
As detailed in the previous section, NMI is one of the popular metrics measuring the performance of unsupervised learning methods and we observe 0.29 units higher NMI in the CNN trained using our proposed framework. In addition to obviously better clustering quality, this metric implies a much higher degree of shared information between the learned clustering and ground truth clustering which in turn shows that the model with a higher NMI is better at extracting the underlying pattern. Further, a loss-less conversion from the rate-based CNN to a spiking network is attempted for the Deep-STDP trained network following the process reported in Ref. \cite{lu2020exploring}. We achieved a similar accuracy (\textbf{0.5662}) in 200 timesteps. Further co-optimization of the SNN accuracy and inference latency can be performed using prior proposals \cite{lu2022neuroevolution}.

\begin{table}[h!]
\centering
\begin{tabular}{|c|c|c|}
\hline
\textbf{Metric} & \textbf{DeepCluster \cite{caron2018deep}} & \textbf{Deep-STDP}\\
\hline
NMI       & 0.5741 & 0.8597\\
FIM       & 0.1890 & 0.5441\\
Acc       & 0.3240 & 0.5696\\
\hline
\end{tabular}
\caption{Evaluation Metrics Comparison}
\label{table: results}
\vspace{-1.2em}
\end{table}

\begin{figure}[htp]
    \centering
    \includegraphics[scale=0.22]{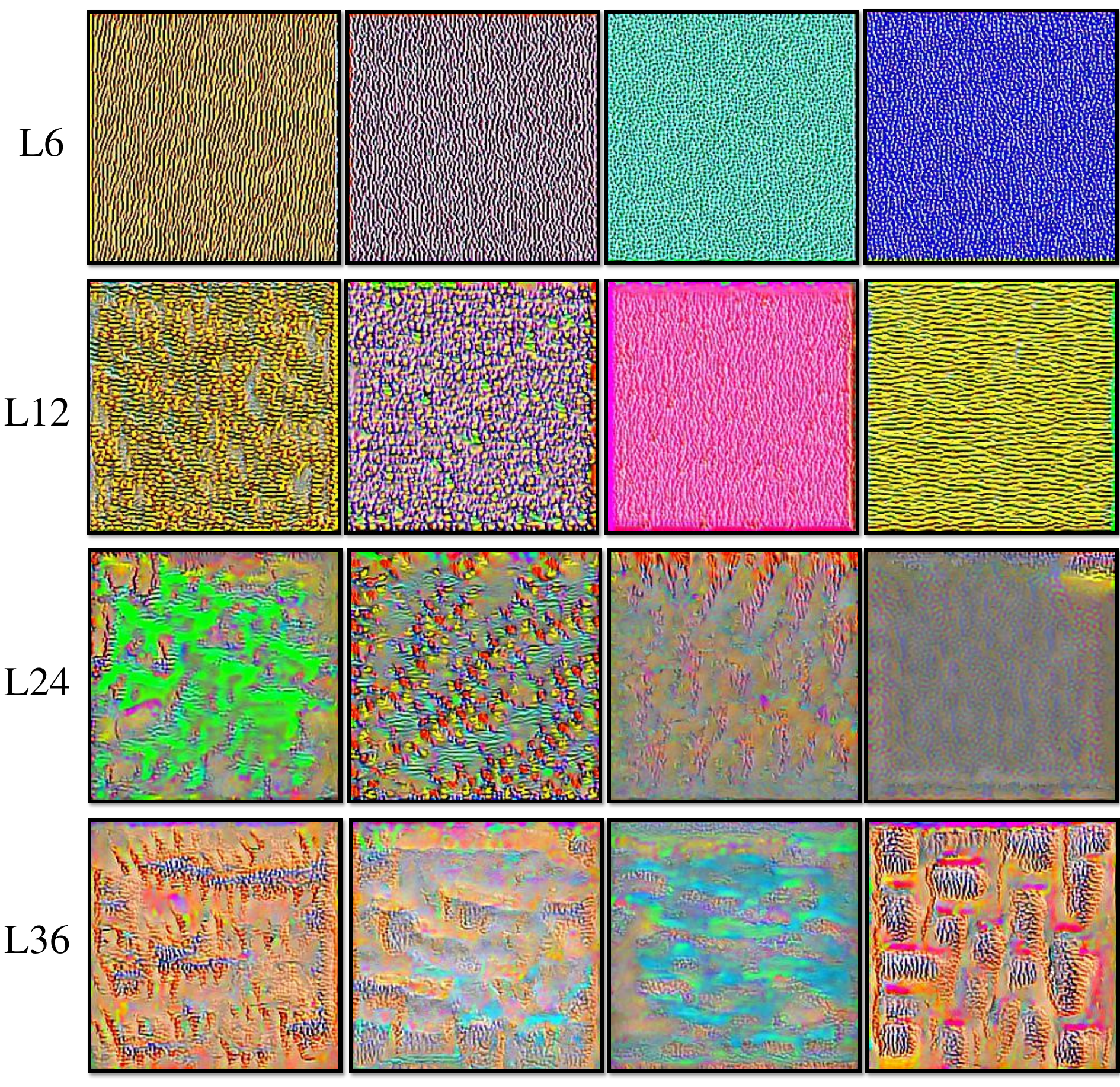}
\caption{Deep-STDP filter activations of Gaussian random noise from layers 6, 12, 24, and 36. We have used the unit-level visualization method proposed in Ref. \cite{erhan2009visualizing}. \label{fig:act}}
\vspace{-2em}
\end{figure}

In addition to training an additional linear layer for numerical performance analysis, we also visualized the convolutional filter activations of the CNN trained using our proposed framework. We can observe from Fig. \ref{fig:act} that the network forms distinct filters specialized for completely different visual patterns in different layers without using any ground truth label information. On the other hand, similar visualization performed on the DeepCluster trained network yielded similar simple patterns in the shallow layers without any complex patterns represented in the deeper layers, further substantiating the efficacy of the Deep-STDP approach.

\subsection{Computational Cost Estimation}
While a detailed system level hardware analysis for the two approaches is outside the scope of this work, we provide motivation for neuromorphic deep clustering by performing a comparative analysis of the computational cost of the two approaches. 
\subsubsection{Cost of k-means Clustering}
To find the new centroid of a particular cluster, the algorithm calculates the averaged center of all the data points assigned to that cluster using the following equation:
\vspace{-1em}
\begin{equation}
c_j = \frac{1}{|C_j|} \sum{x_i \in C_j}
\label{kmeans}
\end{equation}

where, $c_j$ is the averaged coordinates of the $j$-th centroid, $|C_j|$ is the number of data points assigned to that corresponding cluster, and $x_i$ is the $i$-th data point. Subsequently, the algorithm calculates the Euclidean distance between every data point and every centroid and assigns each data point to the cluster with the shortest distance to its centroid. The goal is to solve the optimization problem:
\begin{equation}
    \operatorname*{argmin}_{C} \sum_{j=1}^{k} \sum_{i=1}^{|C_j|} ||x_i - c_j||^2_2
\end{equation}
where, $\operatorname*{argmin}_C$ solves for the optimal centroids and $k$ is the total number of clusters.
The above two calculations will be repeated until convergence is achieved or until a maximum number of iterations is reached. Hence, the number of mathematical operations can be summarized as follows:
\begin{itemize}
    \item \textbf{Clustering Step}: Compute the distance $||x_i - c_j||^2_2$ from every point to every centroid and assign to $k$ clusters 
    \item \textbf{Update Step}: Re-center the centroids in new clusters by averaging over $|C_j|$ for all clusters
    \item \textbf{Repeat} $it$ times
\end{itemize}
To calculate the distance of a point $x_i$ from $c_j$:
\begin{equation}
    ||x_i - c_j||^2_2 = \sqrt{\sum_{m=1}^{d=256}(x_{im} - c_{jm})^2}
\end{equation}
where, $d$ is the number of dimensions in the feature.
Hence, the number of multiplications (the number of squaring operations) in order to calculate the Euclidean distance is:
\begin{equation}
    [k\cdot d] \cdot it \cdot N
\end{equation}
and the number of addition operations involved is:
\begin{equation}
      [k\cdot (2d-1) + d] \cdot it \cdot N
    \label{kmeans_add}
\end{equation}
where, $k$ is the number of clusters, $N$ is the number of training samples, and $it$ is the number of maximum iterations in the $k$-means algorithm. In Eq. \ref{kmeans_add}, the $k \cdot (d-1)$ component arises from the summation of individual distance along each dimension while another $k \cdot d$ component arises from the subtraction operation for distance calculation along each dimension. The last $d$ component arises from updating the new cluster coordinates (which in the worst case will iterate through all data points, see Eq. \ref{kmeans}). Given the cost of float ADD operation is 0.9pJ and float MULT operation is 3.7pJ in 45nm CMOS process \cite{han2015learning}, we estimated the total computational cost in the clustering process for every training epoch to be 14.1mJ (considering $it=20$). Considering 175 epochs of DeepCluster training to reach peak accuracy, the total computational cost is 2467.5mJ. 

\subsubsection{Cost of STDP Clustering}
In the STDP based clustering approach, the computations can be summarized into the following parts: 
\begin{itemize}
    \item \textbf{Feedforward Step}: Integrate input Poisson spike train through the synapses connecting input and excitatory layer
    \item \textbf{Learning Step}: Updating the excitatory layer weights based on pre- and post-synaptic spiking activities
    \item \textbf{Inhibition Step}: Updating the neuron membrane potential based on lateral inhibitory connections
    \item \textbf{Repeat} $T$ times
\end{itemize}

Although multiplication symbols were used in Algo. \ref{alg:pseudo_label}, computation with spike signals can always be reduced to summation operation since the spike magnitude is always `0' or `1' \cite{sengupta2019going}. Further, the addition operation is conditional upon the receipt of spikes, thereby reducing the computation cost by a significant margin for a highly sparse spike train. For instance, the average spiking probability per neuron per timestep in the excitatory layer of the network is only 0.19\%. Hence, the total number of addition operations can be summarized as:
\begin{multline}
[p_{input} \cdot |w^{exc}| + (p_{input} + p_{exc})\cdot |w^{exc}| + p_{exc} \cdot |w^{inh}|]\\
\cdot  T \cdot N 
\label{stdp}
\end{multline}
where, $p_{input},p_{exc}$ are the average (per neuron per timestep averaged over the entire training process) spiking probability of the input and excitatory neuronal layer respectively, $|w^{exc}|$ is the number of synaptic connections between the input and excitatory layer, either $|w_+|$ or $|w_-|$ since the input can be either positive or negative, $|w^{inh}|$ is the total number of inhibitory connections in the network, $T$ is the number of timesteps used for the STDP training process, and $N$ is the number of training samples. 
 
It is worth mentioning here that we primarily focus on the computationally expensive portions of both algorithms for these calculations. In Eq. \ref{stdp}, the $p_{input} \cdot |w^{exc}| $ component arises from the feedforward propagation of input spikes, $(p_{input} + p_{exc})\cdot |w^{exc}|$ component arises from the learning step and  $p_{exc} \cdot |w^{inh}|$ arises from the inhibition step. Therefore, the total computational cost for Deep-STDP per epoch is 55.34mJ and considering 50 epochs of training (iso-accuracy comparison as shown in Fig. \ref{fig:acc}), the total energy consumption is estimated to be 2767.2mJ - comparable to the DeepCluster framework. 

\subsubsection{System Level Cost Comparison:} We note that the STDP based framework does not change the computational load of the clustering framework significantly. However, the computational load at the system level will be also dependent on the computational load for feature extraction in the ConvNet. For instance, Ref. \cite{caron2018deep} mentions a third of the time during a forward pass is attributed to the clustering algorithm while the remaining is attributed to the deep ConvNet feature extraction. Therefore, we expect the Deep-STDP based framework to be significantly more resource efficient than the DeepCluster based approach due to $3.5\times$ reduction in the number of training epochs - equivalently reducing the ConvNet feature extraction computational cost.  


\section{Conclusions}
In conclusion, we proposed an end-to-end hybrid unsupervised framework for training deep CNNs that can be potentially implemented in a neuromorphic setting. We demonstrated significant benefits in terms of accuracy and computational cost by leveraging bio-plausible clustering techniques for deep unsupervised learning of visual features and substantiated our claims by empirical analysis through statistical tools like Fisher Information and Normalized Mutual Information. Our work significantly outperforms prior attempts at scaling bio-inspired learning rules like STDP to deeper networks and complex datasets. Future work can focus on further scaling of the approach and delving deeper into the mathematical underpinnings of the superior performance of STDP as a deep clustering mechanism.  

\section*{Acknowledgments}
This material is based upon work supported in part by the U.S. Department of Energy, Office of Science, Office of Advanced Scientific Computing Research, under Award Number \#DE-SC0021562 and the National Science Foundation grant CCF \#1955815 and by Oracle Cloud credits and related resources provided by the Oracle for Research program.



\begin{thebibliography}{10}
\providecommand{\url}[1]{#1}
\csname url@samestyle\endcsname
\providecommand{\newblock}{\relax}
\providecommand{\bibinfo}[2]{#2}
\providecommand{\BIBentrySTDinterwordspacing}{\spaceskip=0pt\relax}
\providecommand{\BIBentryALTinterwordstretchfactor}{4}
\providecommand{\BIBentryALTinterwordspacing}{\spaceskip=\fontdimen2\font plus
\BIBentryALTinterwordstretchfactor\fontdimen3\font minus
  \fontdimen4\font\relax}
\providecommand{\BIBforeignlanguage}[2]{{%
\expandafter\ifx\csname l@#1\endcsname\relax
\typeout{** WARNING: IEEEtran.bst: No hyphenation pattern has been}%
\typeout{** loaded for the language `#1'. Using the pattern for}%
\typeout{** the default language instead.}%
\else
\language=\csname l@#1\endcsname
\fi
#2}}
\providecommand{\BIBdecl}{\relax}
\BIBdecl




\bibitem{ding2004k}
C.~Ding and X.~He, ``K-means clustering via principal component analysis,'' in
  \emph{Proceedings of the twenty-first international conference on Machine
  learning}, 2004, p.~29.

\bibitem{csurka2004visual}
G.~Csurka, C.~Dance, L.~Fan, J.~Willamowski, and C.~Bray, ``Visual
  categorization with bags of keypoints,'' in \emph{Workshop on statistical
  learning in computer vision, ECCV}, vol.~1, no. 1-22.\hskip 1em plus 0.5em
  minus 0.4em\relax Prague, 2004, pp. 1--2.

\bibitem{caron2018deep}
M.~Caron, P.~Bojanowski, A.~Joulin, and M.~Douze, ``Deep clustering for
  unsupervised learning of visual features,'' in \emph{Proceedings of the
  European conference on computer vision (ECCV)}, 2018, pp. 132--149.

\bibitem{radford2015unsupervised}
A.~Radford, L.~Metz, and S.~Chintala, ``Unsupervised representation learning
  with deep convolutional generative adversarial networks,'' \emph{arXiv
  preprint arXiv:1511.06434}, 2015.

\bibitem{oord2018representation}
A.~v.~d. Oord, Y.~Li, and O.~Vinyals, ``Representation learning with
  contrastive predictive coding,'' \emph{arXiv preprint arXiv:1807.03748},
  2018.

\bibitem{radford2019language}
A.~Radford, J.~Wu, R.~Child, D.~Luan, D.~Amodei, I.~Sutskever \emph{et~al.},
  ``Language models are unsupervised multitask learners,'' \emph{OpenAI blog},
  vol.~1, no.~8, p.~9, 2019.

\bibitem{sengupta2019going}
A.~Sengupta, Y.~Ye, R.~Wang, C.~Liu, and K.~Roy, ``Going deeper in spiking
  neural networks: Vgg and residual architectures,'' \emph{Frontiers in
  neuroscience}, vol.~13, p.~95, 2019.

\bibitem{davies2021advancing}
M.~Davies, A.~Wild, G.~Orchard, Y.~Sandamirskaya, G.~A.~F. Guerra, P.~Joshi,
  P.~Plank, and S.~R. Risbud, ``Advancing neuromorphic computing with loihi: A
  survey of results and outlook,'' \emph{Proceedings of the IEEE}, vol. 109,
  no.~5, pp. 911--934, 2021.

\bibitem{diehl2015unsupervised}
P.~Diehl and M.~Cook, ``Unsupervised learning of digit recognition using
  spike-timing-dependent plasticity,'' \emph{Frontiers in Computational
  Neuroscience}, vol.~9, p.~99, 2015.

\bibitem{saha2021intrinsic}
A.~Saha, A.~Islam, Z.~Zhao, S.~Deng, K.~Ni, and A.~Sengupta, ``Intrinsic
  synaptic plasticity of ferroelectric field effect transistors for online
  learning,'' \emph{Applied Physics Letters}, vol. 119, no.~13, 2021.

\bibitem{frady2020neuromorphic}
E.~P. Frady, G.~Orchard, D.~Florey, N.~Imam, R.~Liu, J.~Mishra, J.~Tse,
  A.~Wild, F.~T. Sommer, and M.~Davies, ``Neuromorphic nearest neighbor search
  using intel's pohoiki springs,'' in \emph{Proceedings of the neuro-inspired
  computational elements workshop}, 2020, pp. 1--10.

\bibitem{bengio2012unsupervised}
Y.~Bengio, A.~C. Courville, and P.~Vincent, ``Unsupervised feature learning and
  deep learning: A review and new perspectives,'' \emph{CoRR, abs/1206.5538},
  vol.~1, no. 2665, p. 2012, 2012.

\bibitem{dike2018unsupervised}
H.~U. Dike, Y.~Zhou, K.~K. Deveerasetty, and Q.~Wu, ``Unsupervised learning
  based on artificial neural network: A review,'' in \emph{2018 IEEE
  International Conference on Cyborg and Bionic Systems (CBS)}.\hskip 1em plus
  0.5em minus 0.4em\relax IEEE, 2018, pp. 322--327.

\bibitem{lloyd1982least}
S.~Lloyd, ``Least squares quantization in pcm,'' \emph{IEEE transactions on
  information theory}, vol.~28, no.~2, pp. 129--137, 1982.

\bibitem{krishna1999genetic}
K.~Krishna and M.~N. Murty, ``Genetic k-means algorithm,'' \emph{IEEE
  Transactions on Systems, Man, and Cybernetics, Part B (Cybernetics)},
  vol.~29, no.~3, pp. 433--439, 1999.

\bibitem{arthur2007k}
D.~Arthur and S.~Vassilvitskii, ``K-means++ the advantages of careful
  seeding,'' in \emph{Proceedings of the eighteenth annual ACM-SIAM symposium
  on Discrete algorithms}, 2007, pp. 1027--1035.

\bibitem{ng2006medical}
H.~Ng, S.~Ong, K.~Foong, P.-S. Goh, and W.~Nowinski, ``Medical image
  segmentation using k-means clustering and improved watershed algorithm,'' in
  \emph{2006 IEEE southwest symposium on image analysis and
  interpretation}.\hskip 1em plus 0.5em minus 0.4em\relax IEEE, 2006, pp.
  61--65.

\bibitem{kim2008recommender}
K.-j. Kim and H.~Ahn, ``A recommender system using ga k-means clustering in an
  online shopping market,'' \emph{Expert systems with applications}, vol.~34,
  no.~2, pp. 1200--1209, 2008.

\bibitem{rumelhart1986learning}
D.~E. Rumelhart, G.~E. Hinton, and R.~J. Williams, ``Learning representations
  by back-propagating errors,'' \emph{nature}, vol. 323, no. 6088, pp.
  533--536, 1986.

\bibitem{hinton2006reducing}
G.~E. Hinton and R.~R. Salakhutdinov, ``Reducing the dimensionality of data
  with neural networks,'' \emph{science}, vol. 313, no. 5786, pp. 504--507,
  2006.

\bibitem{rombach2022high}
R.~Rombach, A.~Blattmann, D.~Lorenz, P.~Esser, and B.~Ommer, ``High-resolution
  image synthesis with latent diffusion models,'' in \emph{Proceedings of the
  IEEE/CVF Conference on Computer Vision and Pattern Recognition}, 2022, pp.
  10\,684--10\,695.

\bibitem{bojanowski2017optimizing}
P.~Bojanowski, A.~Joulin, D.~Lopez-Paz, and A.~Szlam, ``Optimizing the latent
  space of generative networks,'' \emph{arXiv preprint arXiv:1707.05776}, 2017.

\bibitem{kingma2013auto}
D.~P. Kingma and M.~Welling, ``Auto-encoding variational bayes,'' \emph{arXiv
  preprint arXiv:1312.6114}, 2013.

\bibitem{masci2011stacked}
J.~Masci, U.~Meier, D.~Cire{\c{s}}an, and J.~Schmidhuber, ``Stacked
  convolutional auto-encoders for hierarchical feature extraction,'' in
  \emph{Artificial Neural Networks and Machine Learning--ICANN 2011: 21st
  International Conference on Artificial Neural Networks, Espoo, Finland, June
  14-17, 2011, Proceedings, Part I 21}.\hskip 1em plus 0.5em minus 0.4em\relax
  Springer, 2011, pp. 52--59.

\bibitem{diehl2015fast}
P.~U. Diehl, D.~Neil, J.~Binas, M.~Cook, S.-C. Liu, and M.~Pfeiffer,
  ``Fast-classifying, high-accuracy spiking deep networks through weight and
  threshold balancing,'' in \emph{2015 International joint conference on neural
  networks (IJCNN)}.\hskip 1em plus 0.5em minus 0.4em\relax ieee, 2015, pp.
  1--8.

\bibitem{neftci2014event}
E.~Neftci, S.~Das, B.~Pedroni, K.~Kreutz-Delgado, and G.~Cauwenberghs,
  ``Event-driven contrastive divergence for spiking neuromorphic systems,''
  \emph{Frontiers in neuroscience}, vol.~7, p. 272, 2014.

\bibitem{lee2018pretrain}
C.~Lee, P.~Panda, G.~Srinivasan, and K.~Roy, ``Training deep spiking
  convolutional neural networks with stdp-based unsupervised pre-training
  followed by supervised fine-tuning,'' \emph{Frontiers in Neuroscience},
  vol.~12, 2018.

\bibitem{liu2019stdpLearning}
D.~Liu and S.~Yue, ``Event-driven continuous stdp learning with deep structure
  for visual pattern recognition,'' \emph{IEEE Transactions on Cybernetics},
  vol.~49, no.~4, pp. 1377--1390, 2019.

\bibitem{ferre2018unsupervised}
P.~Ferr{\'e}, F.~Mamalet, and S.~J. Thorpe, ``Unsupervised feature learning
  with winner-takes-all based stdp,'' \emph{Frontiers in computational
  neuroscience}, vol.~12, p.~24, 2018.

\bibitem{decolle_kaiser2020synaptic}
J.~Kaiser, H.~Mostafa, and E.~Neftci, ``Synaptic plasticity dynamics for deep
  continuous local learning (DECOLLE),'' \emph{Frontiers in Neuroscience},
  vol.~14, p. 424, 2020.

\bibitem{scellier2017equilibrium}
B.~Scellier and Y.~Bengio, ``Equilibrium propagation: Bridging the gap between energy-based models and backpropagation,'' \emph{Frontiers in Computational Neuroscience}, vol.~11, pp.~24, 2017.

\bibitem{martin2021eqspike}
E.~Martin, M.~Ernoult, J.~Laydevant, S.~Li, D.~Querlioz, T.~Petrisor, and J.~Grollier, ``Eqspike: spike-driven equilibrium propagation for neuromorphic implementations,'' \emph{iScience}, vol.~24, no.~3, 2021.

\bibitem{bal2022sequence}
M.~Bal and A.~Sengupta, ``Sequence learning using equilibrium propagation,''
  \emph{arXiv preprint arXiv:2209.09626}, 2022.

\bibitem{bai2019deep}
S.~Bai, J. Z.~Kolter, and V.~Koltun, ``Deep equilibrium models,'' \emph{Advances in Neural Information Processing Systems}, vol.~32, 2019.

\bibitem{xiao2021training}
M.~Xiao, Q.~Meng, Z.~Zhang, Y.~Wang, and Z.~Lin, ``Training feedback spiking neural networks by implicit differentiation on the equilibrium state,'' \emph{Advances in Neural Information Processing Systems}, vol.~34, pp.~14516--14528, 2021.

\bibitem{bal2023spikingbert}
M.~Bal and A.~Sengupta, ``SpikingBERT: Distilling BERT to train spiking language models using implicit differentiation,'' \emph{arXiv preprint arXiv:2308.10873}, 2023.

\bibitem{noroozi2016unsupervised}
M.~Noroozi and P.~Favaro, ``Unsupervised learning of visual representations by
  solving jigsaw puzzles,'' in \emph{Computer Vision--ECCV 2016: 14th European
  Conference, Amsterdam, The Netherlands, October 11-14, 2016, Proceedings,
  Part VI}.\hskip 1em plus 0.5em minus 0.4em\relax Springer, 2016, pp. 69--84.

\bibitem{midya2019artificial}
R.~Midya, Z.~Wang, S.~Asapu, S.~Joshi, Y.~Li, Y.~Zhuo, W.~Song, H.~Jiang,
  N.~Upadhay, M.~Rao \emph{et~al.}, ``Artificial neural network (ann) to
  spiking neural network (snn) converters based on diffusive memristors,''
  \emph{Advanced Electronic Materials}, vol.~5, no.~9, p. 1900060, 2019.

\bibitem{lu2020exploring}
S.~Lu and A.~Sengupta, ``Exploring the connection between binary and spiking
  neural networks,'' \emph{Frontiers in neuroscience}, vol.~14, 2020.

\bibitem{lu2022neuroevolution}
------, ``Neuroevolution guided hybrid spiking neural network training,''
  \emph{Frontiers in neuroscience}, vol.~16, 2022.

\bibitem{gao2023high}
H.~Gao, J.~He, H.~Wang, T.~Wang, Z.~Zhong, J.~Yu, Y.~Wang, M.~Tian, and C.~Shi,
  ``High-accuracy deep ann-to-snn conversion using quantization-aware training
  framework and calcium-gated bipolar leaky integrate and fire neuron,''
  \emph{Frontiers in Neuroscience}, vol.~17, p. 1141701, 2023.

\bibitem{bellec2018long}
G.~Bellec, D.~Salaj, A.~Subramoney, R.~Legenstein, and W.~Maass, ``Long
  short-term memory and learning-to-learn in networks of spiking neurons,''
  \emph{Advances in neural information processing systems}, vol.~31, 2018.

\bibitem{Rathi2020DIETSNNDI}
N.~Rathi and K.~Roy, ``{DIET-SNN}: Direct input encoding with leakage and
  threshold optimization in deep spiking neural networks,'' \emph{ArXiv}, vol.
  abs/2008.03658, 2020.

\bibitem{caporale2008spike}
N.~Caporale and Y.~Dan, ``Spike timing--dependent plasticity: a hebbian
  learning rule,'' \emph{Annu. Rev. Neurosci.}, vol.~31, pp. 25--46, 2008.

\bibitem{Hazan_2018}
H.~Hazan, D.~J. Saunders, H.~Khan, D.~Patel, D.~T. Sanghavi, H.~T. Siegelmann,
  and R.~Kozma, ``Bindsnet: A machine learning-oriented spiking neural networks
  library in python,'' \emph{Frontiers in Neuroinformatics}, vol.~12, p.~89,
  2018.

\bibitem{deng2012mnist}
L.~Deng, ``The mnist database of handwritten digit images for machine learning
  research,'' \emph{IEEE Signal Processing Magazine}, vol.~29, no.~6, pp.
  141--142, 2012.

\bibitem{deng2009imagenet}
J.~Deng, W.~Dong, R.~Socher, L.-J. Li, K.~Li, and L.~Fei-Fei, ``{ImageNet}: A
  large-scale hierarchical image database,'' in \emph{Computer Vision and
  Pattern Recognition, 2009. CVPR 2009. IEEE Conference on}.\hskip 1em plus
  0.5em minus 0.4em\relax IEEE, 2009, pp. 248--255.

\bibitem{zhang2016colorful}
R.~Zhang, P.~Isola, and A.~A. Efros, ``Colorful image colorization,'' in
  \emph{Computer Vision--ECCV 2016: 14th European Conference, Amsterdam, The
  Netherlands, October 11-14, 2016, Proceedings, Part III 14}.\hskip 1em plus
  0.5em minus 0.4em\relax Springer, 2016, pp. 649--666.

\bibitem{amari2000methods}
S.-i. Amari and H.~Nagaoka, \emph{Methods of information geometry}.\hskip 1em
  plus 0.5em minus 0.4em\relax American Mathematical Soc., 2000, vol. 191.

\bibitem{karakida2019universal}
R.~Karakida, S.~Akaho, and S.-i. Amari, ``Universal statistics of fisher
  information in deep neural networks: Mean field approach,'' in \emph{The 22nd
  International Conference on Artificial Intelligence and Statistics}.\hskip
  1em plus 0.5em minus 0.4em\relax PMLR, 2019, pp. 1032--1041.

\bibitem{kim2022exploring}
Y.~Kim, Y.~Li, H.~Park, Y.~Venkatesha, A.~Hambitzer, and P.~Panda, ``Exploring
  temporal information dynamics in spiking neural networks,'' \emph{arXiv
  preprint arXiv:2211.14406}, 2022.

\bibitem{erhan2009visualizing}
D.~Erhan, Y.~Bengio, A.~Courville, and P.~Vincent, ``Visualizing higher-layer
  features of a deep network,'' \emph{University of Montreal}, vol. 1341,
  no.~3, p.~1, 2009.

\bibitem{han2015learning}
S.~Han, J.~Pool, J.~Tran, and W.~Dally, ``Learning both weights and connections
  for efficient neural network,'' \emph{Advances in neural information
  processing systems}, vol.~28, 2015.

\end{thebibliography}



\end{document}